\documentclass[conference]{IEEEtran}
\IEEEoverridecommandlockouts
\usepackage{cite}
\usepackage{amsmath,amssymb,amsfonts}
\usepackage{algorithmic}
\usepackage{graphicx}
\usepackage[caption=false]{subfig}
\usepackage{textcomp}
\usepackage{color}
\usepackage{multirow}
\usepackage{hyperref}
\usepackage[dvipsnames]{xcolor}

\def\BibTeX{{\rm B\kern-.05em{\sc i\kern-.025em b}\kern-.08em
 T\kern-.1667em\lower.7ex\hbox{E}\kern-.125emX}}

\begin{document}

\title{Robust and Transferable Anomaly Detection in Log Data using Pre-Trained Language Models\\
}
\author{\IEEEauthorblockN{Harold Ott\IEEEauthorrefmark{1},\
Jasmin Bogatinovski\IEEEauthorrefmark{1},
Alexander Acker,
Sasho Nedelkoski,
Odej Kao 
}
\IEEEauthorblockA{Distributed and Operating Systems, 
TU Berlin, Berlin, Germany\\ Email:  \{jasmin.bogatinovski, alexander.acker, nedelkoski, odej.kao\}@tu-berlin.de}
\IEEEauthorblockA{\IEEEauthorrefmark{1}Equal contribution.}
}
\maketitle

\begin{abstract}
Anomalies or failures in large computer systems, such as the cloud, have an impact on a large number of users that communicate, compute, and store information. 
Therefore, timely and accurate anomaly detection is necessary for reliability, security, safe operation, and mitigation of losses in these increasingly important systems. Recently, the evolution of the software industry opens up several problems that need to be tackled including (1) addressing the software evolution due software upgrades, and (2) solving the cold-start problem, where data from the system of interest is not available.
In this paper, we propose a framework for anomaly detection in log data, as a major troubleshooting source of system information. To that end, we utilize pre-trained general-purpose language models to preserve the semantics of log messages and map them into log vector embeddings. The key idea is that these representations for the logs are robust and less invariant to changes in the logs, and therefore, result in a better generalization of the anomaly detection models. We perform several experiments on a cloud dataset evaluating different language models for obtaining numerical log representations such as BERT, GPT-2, and XL. The robustness is evaluated by gradually altering log messages, to simulate a change in semantics. Our results show that the proposed approach achieves high performance and robustness, which opens up possibilities for future research in this direction. 
\end{abstract}

\begin{IEEEkeywords}
anomaly detection, log analysis, deep learning, language models, transfer learning
\end{IEEEkeywords}

\section{Introduction}

Modern computer systems such as cloud platforms are a combination of complex multi-layered software and hardware. 
The complexity implies high maintenance overhead for the operators of these systems, making the manual operation cumbersome. In extreme cases, where system anomalies or failures happen, it can lead to SLA violations. Large service providers are aware of such cases and make the automation operation and maintenance tasks a priority.

Recently, a plethora of methods were introduced to automate and provide scalable AI-driven solutions to perform a range of operational tasks including anomaly detection and failure analysis~\cite{bogatinovski2021multi,nedelkoski2020selfattentive,zhu2019tools}. In the foundation of these methods are the system data. Although there are various data sources describing system behaviour, system logs are an omnipresent data source~\cite{zhu2019tools,du2017deeplog}. They are one of the most used data sources for troubleshooting. 

The evolution of the software industry opens up several problems that need to be tackled. The detection of the abnormal behaviour of the system is one of them. When considering the anomaly detection models from log data,  two of the most important challenges are (1) addressing the software evolution due software upgrades, and (2) solving the cold-start problem~\cite{zhang2019robust}. In both cases, anomaly detection models have to be dynamically optimized and adapted to the new setting. Exposing the underlying properties of the log messages in a system-agnostic manner (e.g. semantics, length, etc.) arises as an important requirement from the anomaly detection methods utilizing system logs.

On the contrary, many of the existing approaches are based on the invariant assumption, i.e. log templates never change. Furthermore, they rely on the assumption capturing all possible variations of log messages. 
Approaches, such as matching certain keywords (e.g. "error"), constructing a black-list of log events or anomalous matching regular expressions, are infertile under the circumstances of constant system's evolution. They usually lead to many unnecessary alarms, a problem known as alarm fatigue. 

To mitigate the drawbacks of the invariant assumption, we propose an anomaly detection framework capable of preserving the shared properties between the log messages. More specifically, we utilized transfer learning and deep language modeling to learn a robust, context-aware representation of the log messages. Whenever a new log line is introduced, the framework assigns numerical-vector representation to it utilizing prior information from all the previously presented log messages. As such it is effective in reducing the cold-start problem the anomaly detection model is facing after an update. Through time, the framework reuses the accumulated knowledge for the log messages to improve the performance and the underlying representation. In a nutshell, the framework provides a mechanism to transfer knowledge from previous log messages and automatically detect anomalies in logs affected by pre-processing noise and changes of log events by updates of the underlying software.

The contributions of this work are summarized as follows.
\begin{enumerate}
    \item A general framework for learning context and semantic-aware numerical log vector representations suitable for anomaly detection.
    \item Comparison of three semantic-level general-purpose language embedding models for anomaly detection.
    \item Comparison of two learning objectives for anomaly detection utilizing general language models.
    \item Robust model transfer approach for reduction of the false positive rate after software update. 
    \item We provide a publicly available implementation of the method and the datasets.\footnote{{https://github.com/haraldott/anomaly\_detection\_main}}
\end{enumerate}

The remaining of the paper is structured as follows. In section~\ref{related_work}, we provide the related work for anomaly detection in log data. In sections \ref{methods}, we present the preliminary the proposed framework. Section \ref{evaluation} summarizes the evaluation of the different language models, learning objective as well as model transfer. Section~\ref{conclusion} concludes the paper.

\section{Related Work}\label{related_work}
As a major data type for the system behaviour, the literature recognizes sustainable utilization of the log data for anomaly detection in both the industry and academia~\cite{du2017deeplog,xu2009detecting,zhang2019robust,zhu2019tools,nedelkoski2020selfsupervised}. The work on anomaly detection from log data follows two general directions: supervised and unsupervised methods. In this work, our focus is on unsupervised learning approaches.


The unsupervised approaches have greater practical relevance, because labeling of log messages is an expensive procedure.
There is a number of approaches that have been developed using the log event count within a certain window to transform log messages to numerical representations. Xu et al.~\cite{xu2009detecting} proposed using the Principal Component Analysis (PCA) method on such vectors. It follows a standard machine learning techniques of investigation of the second norm of the lower principle components to decide if the log is normal or an anomaly.  


There are several works for log anomaly detection that utilize deep learning approaches. 
For example, Zhang et al.~\cite{Zhang2016AutomatedIS} used LSTM to predict subsequent log events based on a window of preceding events. The ability to correctly predicting the next event is used to determine anomalous events.
DeepLog~\cite{du2017deeplog} is utilizing a similar method. It is claimed that robustness to novel events is achieved by a synonym/antonym database that is used to generate auxiliary samples. Vijayakumar et al.~\cite{Vinayakumar2017LongSM} trained a stacked-LSTM to model the operation log samples of normal and anomalous events. However, the input to the unsupervised methods is a one-hot vector of logs representing the indices of the log templates. Therefore, it cannot cope with newly appearing log events.

Several studies \cite{zhang2019robust} have leveraged NLP techniques to analyze log data based on the idea that log is a natural language sequence. Those works are utilizing word embeddings which are later averaged in order to represent the full log message. Non-learnable aggregation is a heuristic that often does not hold when going from words to sentences~\cite{mikolov2013distributed}. Different from all the above methods, we utilize state-of-the-art language models for obtaining numerical representations for log messages. It enables using end-to-end trainable vector representations that can be used in various recurrent networks e.g. Bi-LSTM~\cite{huang2015bidirectional} for anomaly detection. The log representations are robust to semantic-invariant changes of the log messages, providing good generalization.

\begin{figure}[b]
	\centering
	\includegraphics[width=0.45\textwidth]{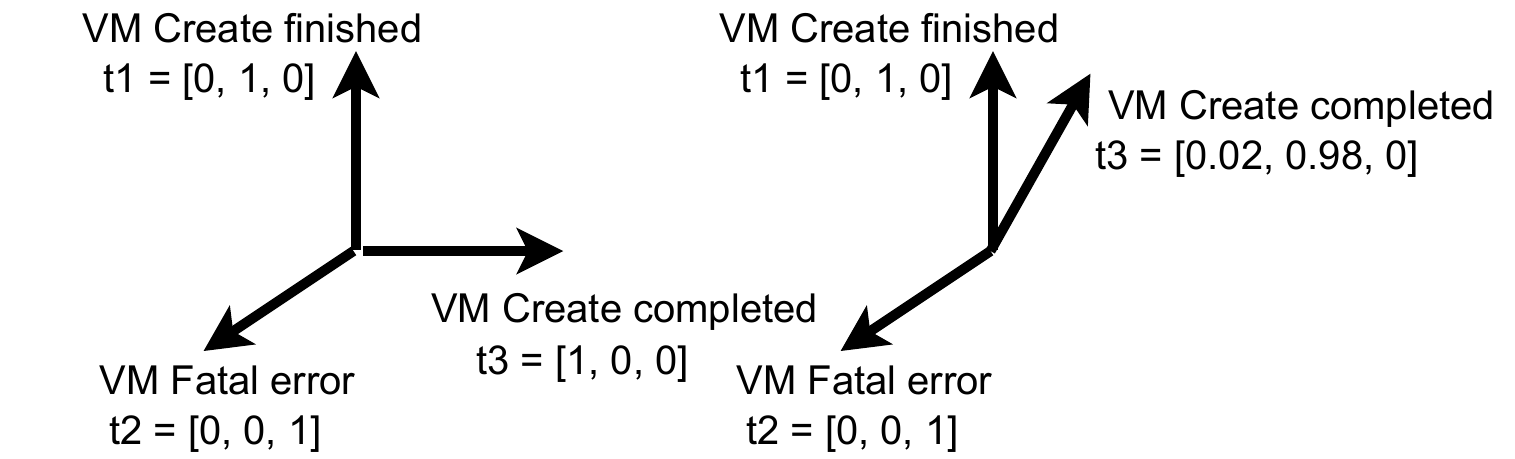}
	\caption{Log vector representation using invariant embeddings (left) and semantically-aware embeddings (right). }
	\label{fig:vectors}
\end{figure}

\section{Robust Log Anomaly Detection~\label{methods}}
\begin{figure*}[t]
	\centering
	\includegraphics[width=0.7\textwidth]{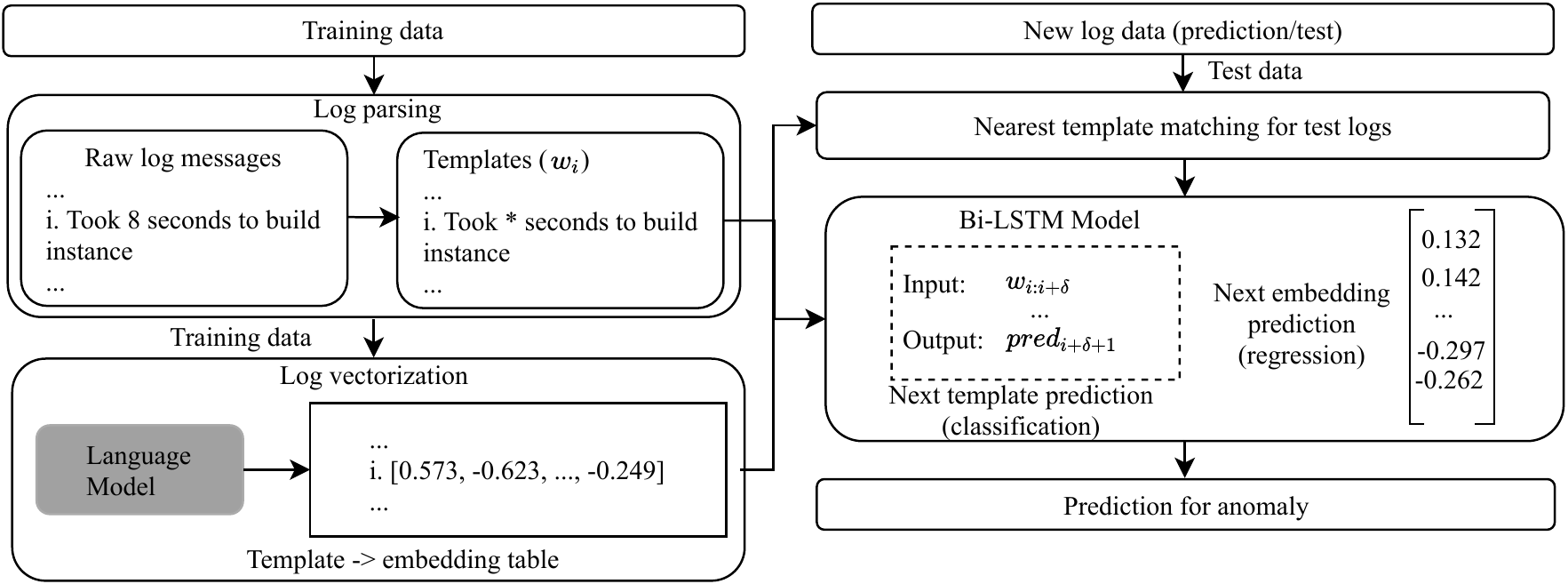}
	\caption{Overview of the framework utilizing sentence level pre-trained language models.}
	\label{fig:overall_system}
\end{figure*}

The architecture of the framework is presented in Figure~\ref{fig:overall_system}.
It is composed of two phases, offline training phase and online anomaly detection.

\textbf{The training phase} is composed of the following steps. First, the raw log messages from the system are preprocessed. This includes a transformation of the log into a template and variable part (e.g., \verb!VM Creation took 8 seconds!; template: "\verb!VM Creation took * seconds!", variables: [8]). Each of the templates is then transformed into a numerical vector using language models such as BERT, GPT, and XL~\cite{devlin2018bert,radford2019language,yang2019xlnet}. Utilizing the pre-trained embeddings from these general-purpose models aims to capture the semantic properties the log messages, important for generalization over different data~\cite{zhang2019robust}. 
In the second step, we chain the embedding vectors through time in a recurrent neural network (Bi-LSTM~\cite{huang2015bidirectional}) that learns the normal system behaviour. It is then utilized for anomaly detection by detecting deviations from the expected system behaviour. It enables robust detection of sequential anomalies. Important to note is that the neural network is trained on pre-trained numerical representations, therefore, it largely facilitates the transfer of the learned model to new log data that can appear due to software upgrades or due cold-start problems. 

In \textbf{the prediction phase}, the log messages are transformed into log vectors via the same preprocessing steps as in training. Then, the sequences of such log vectors are provided as input to the anomaly detection model. The prediction from the sequential model is utilized to decide if the input sequence is normal or not. In the following, we describe each of the parts in detail.

\subsection{Log preprocessing and log vectorization \label{sec:pre_processing}}
The raw log messages generated by the systems are noisy with semi-structured form. To structure them and to obtain the information from the logs needed for the anomaly detection model, they require to be parsed~\cite{zhu2019tools}. Log parsing provides a mapping function of the raw log messages into log templates e.g log instruction in the source code. In this work, we adopt Drain~\cite{he2017drain}, due to its speed and efficiency.


Next, the log templates are transformed into numerical vectors. Formally, we write a log vector (embedding) as $w_i \in R^{d}$, where $d$ is the size of the vector embedding. The goal of the log vectorization is to preserve important properties of log messages and distinct normal against anomaly log messages.

To better illustrate the importance of the log vectors, in Figure~\ref{fig:vectors}, we provide a visual comparison between normal and anomalous log messages when standard one-hot encoding is utilized against vector embeddings obtained from pre-trained methods. Improvements in the log vectorization translate to improvements in the robustness and generalization of the anomaly detection models.

To that end, we formalize two properties that a log vector embedding should poses. (1) Distinguishable: the log vectors should represent semantic differences between the log messages. For example, \verb!VM! \verb!Create! \verb!finished! and \verb!VM! \verb!Fatal! \verb!error! are templates with different semantics, even though they share the same words (instance) and synonyms (terminating, deleting). (2) Tolerance: the embeddings should represent the similarity between different templates with the same or very similar semantics. For example, \verb!VM! \verb!Create! \verb!finished! is semantically very similar to \verb!VM! \verb!Create! \verb!completed!.

To preserve both properties, we refer to the natural language models, where these properties are one of the major parts of research. Exploiting general-purpose language models, which are pre-trained on large corpora of texts (e.g., Wikipedia) enables preserving of general textual structures. We focus on utilizing sentence-level embeddings, in contrary to word level embeddings. Sentence level embedding provide direct and efficient mapping from log message to log vector, without any intermediate steps (e.g. averaging of word vectors).



\subsection{Bi-LSTM for Sequential Anomaly Detection\label{sec:prediction_model}}
Once obtained, the log vectors are grouped (by timestamp) into sequences of size $\delta$ (window size), i.e., the sequences are formed of consecutive log vector embeddings, $w_{i}:w_{i+\delta}$. We define two learning tasks which are utilized to learn the normal sequences of log messages: (1)Prediction of the log template as a class (classification, via minimization of the cross-entropy loss), and (2) prediction of the log vector (regression, via minimization of the mean squared error), of the log message that appears at the next position in the sequence $W_{i+\delta+1}$, given the $w_{i}:w_{i+\delta}$ sequence as input. 

\begin{figure}[b]
	\centering	
	\includegraphics[width=0.75\columnwidth]{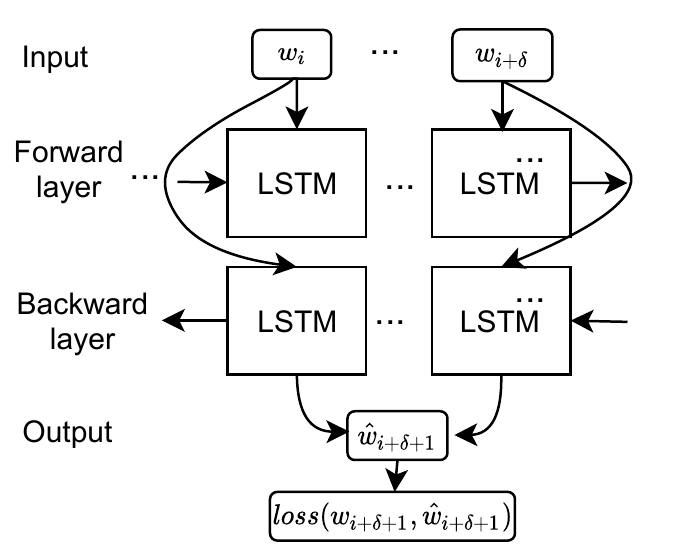}
	\caption{Unfolded Bi-LSTM model used for anomaly detection of the embedding sequence.}
	\label{fig:lstm_model}
\end{figure}

Figure~\ref{fig:lstm_model} depicts the overview of the Bi-LSTM model used to optimize the objectives~\cite{huang2015bidirectional}. The input data is passed to the forward and backward layers of the Bi-LSTM. We selected this model for learning the sequences as it offers a two-sided view and improved properties for sequence learning, in comparison to the single LSTM networks.

The output of the Bi-LSTM network is an abstract numerical representation of the input sequence, which is then utilized for optimizing the objective. The subsequent two linear layers are applying a transformation to acquire the desired dimensions, i.e., $d$ for regression and $n$ (number of classes) for classification. Finally, an activation function $f$ is applied to the output of the last linear layer. We use cross-entropy for classification and mean squared error for regression.

\textbf{Anomaly Detection using Multi-Class Classification}. For this learning objective, we used all available log templates as a target class (total of $n$). The training is performed on the assumption that the data contains an abundance of normal log messages, while in the prediction phase, the input data contains normal and anomalous log templates. 

One major issue in this setup is that of the "close-world" classification objective requires apriori knowledge of all log templates. However, during prediction, it is expected that novel log templates will emerge. To address the absence of all templates at the prediction phase, we apply a nearest template matching procedure to mitigate this limitation. 

In the template matching procedure, we calculate the distance between the embedding of the novel template and all of the known target embeddings. The novel template is assigned the class target that has the smallest distances to the known target templates. To prevent matching on arbitrary novelties, a parameter $maximal\ distance$ is introduced. When the minimal distance to the template to the set of known templates greater then some the $maximal\ distance$, the novel template is discarded and anomaly label is directly assigned. The matching process is applied on $w^{i}_{i+\delta+1}$ in order to obtain $t_{i+\delta+1}$.



After the matching process, the model predicts a probability distribution $Pr[t_{i+\delta+1}|w_{i:i+\delta}]$ = $(p(t) | \forall t \in \mathbb{T})$.
It described the probability of a template $t \in \mathbb{T}$ to occur as a successor of templates $w_{i:i+\delta}$. Due to the noise in the sequential appearing of the templates, we consider the top-$k$ (out of $|T|$) templates with the highest probabilities to appear as relevant as the next template. If the actual template class $t_{i+\delta+1}$ is within the $top-k$ predictions with the highest probability, we consider is as normal. Otherwise, it is labeled as an anomaly.



\textbf{Anomaly Detection using Log Vector Regression}. For the regression learning objective, the neural network is trained to minimize the mean squared error (MSE). The input of the network is a sequence of vector embeddings for the templates, while the corresponding target value for the sequence is the vector embedding of the next template. Compared to classification, the log vector embeddings for regression are always obtainable.

After the model is trained, the parameters for the anomaly detection models are calculated. The regression anomaly detection module has as a parameter the $q-th$ percentile of the squared error for the training samples. The mean squared error of every target for each training sequence template at position $i+s+1$ 
and the neural network's predicted template vector, is computed. Afterwards, the $q$-th percentile of the agglomerated loss values of the training dataset is computed. To detect anomaly, when novel sample from test dataset is introduced, the squared error between the predicted template and the vector embedding of the nearest matched template. The system will then mark every log event whose embedding loss value is above the calculated $q$-th percentile as an anomaly and normal, otherwise. 

\subsection{Model Transfer~\label{sec:transferlearning}}
Utilizing pre-trained general-purpose language models for extracting log representations and training the Bi-LSTM model allows the transfer of the model to new unseen logs. The model transfer is achieved in the following way.  

Let dataset $A$ be the training dataset from already known log messages, and dataset $B$ be a dataset from an updated or new service or system. After the preprocessing, the model is trained on the dataset $A$. Then, the following steps are executed. First, every log event of dataset $B$ is mapped to the nearest neighbour of dataset $A$, i.e. the embedding with the shortest cosine distance. In the case of classification, it gets assigned the same class target. Second, a few-shot training on dataset $B$ will be executed. Finally, with the adjusted model on training dataset $B$, the prediction phase on a test dataset $B$ is executed as previously described for the classification and regression learning objectives. 

The initial training on dataset $A$ preserves semantic and contextual information from previous log messages. The few-show training on dataset $B$ allows the model to adapt to the specifics of the dataset $B$ and improve the results on anomaly detection. 


\section{Evaluation\label{evaluation}}
To demonstrate the usefulness of our framework for anomaly detection and transferability of the models from different software deployments we conducted two evaluation experiments. In the first experimental scenario, we investigate how effective are the representations from sentence-level language models for anomaly detection on 1) ground truth anomalies and 2) synthetic anomalies obtained via log alteration. In the second experimental scenario, we evaluate the transferability of the models during software updates.

\subsection{Log Datasets~\label{sec:ad_one_ds_result}}
For our experiments, we utilize the CloudLab OpenStack log dataset available at the Loghub~\cite{du2017deeplog}. It is composed of two sets of experiments. During the first set of experiments, the Openstack instances were created and their runtime was monitored. The second set of experiments is similar to the former, however, occasionally anomalies were injected. The first dataset, we refer to as a normal dataset while the second one as anomalous dataset.
Furthermore, to evaluate the framework, we additionally manipulated the normal dataset and created two additional test sets described in the following.


\begin{table*}[!t]
\caption{Comparison of the sentance-level language embedding models for the task of anomaly detection.}
\label{tab:languages_eval}
\centering
\begin{tabular}{|c|c|l|l|l|l|l|l|l|l|l|}
\hline
\multirow{2}{*}{learning  objective}                            & \multirow{2}{*}{\begin{tabular}[c]{@{}l@{}}type of\\ experiment\end{tabular}} & \multicolumn{3}{c|}{Precision score} & \multicolumn{3}{c|}{Recall score} & \multicolumn{3}{c|}{F1 score} \\ \cline{3-11} 
                                                  &                                                                             & GPT-2     & XL       & BERT    & GPT-2    & XL      & BERT   & GPT-2   & XL    & BERT  \\ \hline
\multicolumn{1}{|c|}{\multirow{3}{*}{regression}} & semantic                                                                    & 0.88      & 0.21     & 0.43    & 1.00     & 0.63    & 1.00   & 0.94    & 0.31  & 0.56  \\ \cline{2-11} 
\multicolumn{1}{|c|}{}                            & sequential                                                                  & 0.79      & 0.32     & 0.49    & 1.00     & 0.61    & 1.00   & 0.87    & 0.42  & 0.66  \\ \cline{2-11} 
\hline
\multirow{3}{*}{classification}                   & semantic                                                                      & 0.24      & 0.26     & 0.37    & 0.70     & 1.00    & 1.00   & 0.36    & 0.41  & 0.54  \\ \cline{2-11} 
                                                  & sequantial                                                                      & 0.31      & 0.36     & 0.5     & 0.70     & 1.00    & 1.00   & 0.43    & 0.53  & 0.67  \\ \cline{2-11} 
                                                \hline
\end{tabular}
\end{table*}

\textit{Log alteration\label{sec:logs_alteration}}. To evaluate the feasibility of sentence-level based embeddings for anomaly detection in log data we augmented our data with a synthetic dataset. We refer to this data as log alteration data. We identified two points of alteration in the log messages; semantic and contextual alteration. The alterations are applied to normal data. Therefore, the overall anomaly detection model should be robust against these alterations.
Classifying suchlike altered log messages as anomalies are considered as false alarms. For both, the semantic and structural changes we identified 3 types of alteration, namely: deletion, swap and imputation. 

For the semantic changes, we assume a log event to contains $n$ tokens originating from the normal data. Deletion operation involves deleting of $l$ randomly selected words in the log message. Swap operation involves, replacing $l$ tokens with a random token. Imputation operation involves imputing $l$ words at a random position of original log event. The parameter $l$ controls the intensity of the alternation. It is expected that log events with higher alternation intensity have a higher probability to be detected as anomalies compared to events that were altered with lower intensity. 

For the structural changes, we assume a log sequence to contains $m$ log templates originating from the normal data. Deletion operation involves deleting of $l$ random log events from the sequence. Swap operation involves, replacing the $k$ templates appearing after randomly selected index $i$, to a randomly chosen index $j$. For the indices the inequality $j<i$ holds. Imputation operation for sequences involves selecting index at position $i$ and repeating it $l$ times consecutively. The parameter $l$ controls the number of imputations. It is expected that log events with higher alternation intensity have a higher probability to be detected as anomalies compared to events that were altered with lower intensity.

\textit{Augmentation to Simulate a Different Dataset\label{sec:transfer_learning_setup}}. 
Since the software is often updated and thus changed constantly by developers, log statements are also subject to change. To simulate the evolution of the system, we construed an artificial dataset that simulates changed log messages. We constructed two datasets, we refer to as dataset A and B, in the following manner. We start with the normal data we refer to dataset A. Firstly, we randomly sample $p\%$ of the logs in A. Secondly, the sampled log lines are altered using the three semantic alteration techniques with additional word augmentation. The alteration parameters are set to random values in the range 5-100 $\%$ of the range of allowed values for altering parameters. This allows simulating different dataset.  We refer to this altered dataset as a dataset B. Finally, we create two versions of the dataset B. If the alteration is not severe (e.g. 20\% of the log messages is changed) the dataset is referred to as \textit{B-similar}, otherwise, the dataset is referred to as \textit{B-different}.
The datasets A and B are used for transferring the contextual and semantic accumulated knowledge in the following way. The model is trained on this dataset A for $e$ epochs (60 in our study). Then part of the dataset B is used to conduct few-shot training. The final evaluation is done on the task of anomaly detection in the second part of dataset B. 



\subsection{Semantic-level language embedding evaluation\label{sec:evaluation}}
This section presents the evaluation of the results. We first evaluate the sentence-level embeddings capability of the different language models for anomaly detection independent on the learning task. Namely, we compare BERT, XL-Transformers and GPT-2 on the regression-based approach and the classification-based approach for anomaly detection. 
Afterwards, the results of the evaluation using the model transfer learning approach are presented.

\subsubsection{Regression-based anomaly detection.\label{sec:results-regression}}
\tablename~\ref{tab:languages_eval} enlists the results from the comparison of the three language models on the task of anomaly detection.
We divided the experiments into two subsets according to the type of alteration. Semantic alteration is related to the semantic changes of the log messages, while the sequential alteration is related to sequential alteration, described previously.
For the semantically alerted log messages, GPT-2 yields better results compared to BERT and XL-Transformers with regards to all metrics. For the sequential altering of the log messages, there is a small drop of F1-score and precision for the GPT-2 embeddings. However, the same metrics increase for BERT and XL embedding. The results from both scenarios imply that GPT-2 and BERT embeddings are more robust when either the semantic or sequential changes are considered.

\subsubsection{Classification \label{sec:results-classification}}
When considering the classification task we conducted the two separate results as in the case of regression. 
For semantically alerted log messages the scores are reversed. More specifically, BERT is showing the best results, followed by XL and GPT-2. The same pattern appears when considering the sequential learning scenario. 
General comparison of the scores between the regression and classification tasks shows that GPT-2 embeddings are highly affected by the optimization objective.  The definition of the problem as a classification task is favourable when considering structural and sequential changes.

\begin{table*}[ht]
\caption{Evaluation results for the model transfer after software updates. Percentage of altered log messages is p=15$\%$.}
\label{tab:transferModelEval}
\centering
\begin{tabular}{|l|l|l|l|l|l|l|l|l|l|l|}
\hline
\multirow{2}{*}{learning objective}                            & \multirow{2}{*}{\begin{tabular}[c]{@{}l@{}}type of\\ experiment\end{tabular}} & \multicolumn{3}{c|}{Precision score} & \multicolumn{3}{c|}{Recall score} & \multicolumn{3}{c|}{F1 score} \\ \cline{3-11} 
                                                  &                                                                             & GPT-2     & XL       & BERT    & GPT-2    & XL      & BERT   & GPT-2   & XL    & BERT  \\ \hline
\multicolumn{1}{|c|}{\multirow{2}{*}{regression}} & B-similar                                                                     & 0.23      & 0.45     & 0.58    & 0.05     & 0.7     & 0.7    & 0.08    & 0.55  & 0.63  \\ \cline{2-11} 
\multicolumn{1}{|c|}{}                            & B-different                                                                   & 0.94      & 0.18     & 0.52    & 1.00     & 0.47    & 1.00   & 0.97    & 0.26  & 0.68  \\ \hline
\multirow{2}{*}{classification}                   & B-similar                                                                     & 0.27      & 0.53     & 0.61    & 1.00     & 1.00    & 1.00   & 0.43    & 0.69  & 0.75  \\ \cline{2-11} 
                                                  & B-different                                                                   & 0.09      & 0.23     & 0.68    & 1.00     & 1.00    & 1.00   & 0.17    & 0.38  & 0.81  \\ \hline
\end{tabular}
\end{table*}


\subsection{Model Transfer Evaluation \label{sec:results_transfer_regression}}
For the evaluation of model transfer we conducted two experiments, for both the regression- and classification learning objectives on the task of anomaly detection. The results are listed in \tablename~\ref{tab:transferModelEval}.

For the regression learning objective when considering the large alteration, it can be seen that the both GPT-2 and BERT are performing well. However, when considering the small alteration, although BERT embeddings still retain high score, the GPT-2 tends to produce weaker results. On contrary, while being good performing method on the task of similar log messages, XL-Transformers fails when the changes of the log messages are not drastic. 

On the classification learning objective, when considering both large and small alterations, the model utilizing BERT tends to outperform the remaining two. Comparing the XL-Transformers and GPT-2, it can be observed that the former outperforms the latter. Comparing the results alongside the learning objectives, it can be observed that the classification problem definition, slightly outperforms the definition of the problem as a regression task.



\subsection{Discussion}
The good results from both the classification and regression learning objectives show that the framework is useful for anomaly detection in setting where the data is evolving through time. When evaluating the different forms of alteration of the log messages and sequences of log messages BERT, as a general-purpose language model on sentence-level embeddings, shows to perform more consistently and robustly across the two learning objectives. It is followed by XL-Transformers and GPT-2 accordingly. GPT-2 shows strong results in experiment type for regression learning objective, but not as competitive for classification learning objective. Similar observations can be made for model transfer in settings where there are both small and large changes in the log messages. 

Comparison of the different learning objectives shows that the definition of the learning task as a classification problem can produce better results compared to it defined as a regression problem. This is an interesting result from this study.

The plug-and-play strategy allows for testing different language models. As seen by the results, the different language models can highly influence the quality of the results for anomaly detection, with different word embeddings having strengths and weaknesses in different categories. Improving the NLP language models via increasing the number of parameters e.g.~\cite{brown2020language} will result in even better performance.


\section{Conclusion} \label{conclusion}
This paper addresses the problem of log anomaly detection in large computer systems.  
We addressed the generalization problem for anomaly detection on unseen logs by introducing a plug-and-play framework that utilizes pre-trained language models for obtaining numerical, semantically aware embeddings for log events.  Bi-LSTM neural network is used as a method for exploiting contextual properties of log messages in the task of anomaly detection. Empirically, we show that the proposed approach is robust to alteration in the log messages -- scenarios frequently occurring in practice due to software updates and deploying new services or systems. 
The results show that the framework achieves high performance using state-of-the-art sentence-level language models. 
Furthermore, we show that not every representation is equally useful for anomaly detection. Some of the language models fail to generate log representations that can be separated by a learned decision boundary. The underlying learning objective is also very important to obtain good results in the task of anomaly detection. The proposed approach opens new potential for anomaly detection not just from log data, but from other sources that have the notion of a distributed representation of an event e.g., distributed tracing data. We believe that the method will motivate further research in the direction of development of pre-trained language models on log data. This would enhance the log representations, and thus, improve the performance of the anomaly detection methods.





\bibliographystyle{IEEEtran}
\bibliography{shorter_version}
\end{document}